\pdfoutput=1
\documentclass[11pt]{article}
\usepackage[final]{acl}

\usepackage{times}
\usepackage{latexsym}
\usepackage[tone]{tipa}
\usepackage{tikz}
\usepackage{makecell}
\usepackage{rotating}
\usepackage{amssymb}
\usepackage{pifont} 
\usepackage{tabularray} 
\usepackage{graphicx}
\usepackage{array}
\usepackage{fdsymbol}
\usepackage{cleveref}
\usepackage{booktabs}
\usepackage{amsmath}
\usepackage{multirow}
\usepackage{xspace}
\usepackage{subcaption}
\usepackage{enumitem}

\usepackage{tablefootnote}
\usepackage{threeparttable}
\usepackage{rotating}

\usepackage{lipsum}
\usepackage{tcolorbox}
\usepackage{xcolor}

\usepackage[T1]{fontenc}

\usepackage[utf8]{inputenc}

\usepackage{microtype}
\usepackage{inconsolata}

\usepackage{twemojis}

\usepackage{tcolorbox}

\usepackage{tabularray}
\usepackage{cellspace}
\newcommand\cincludegraphics[2][]{\raisebox{-0.3\height}{\includegraphics[#1]{#2}}}
\usepackage{setspace}

\newcommand{\ttipa}[1]{\texttt{\textipa{#1}}}

\newcommand{\ipachildes}[0]{\textsc{IPA CHILDES}\xspace}

\title{BabyLM's First Words:\\ Word Segmentation as a Phonological Probing Task}

\author{
    {\bf Z\'{e}bulon Goriely} \texttwemoji{orange} ~~~~~ 
    {\bf Paula Buttery} \texttwemoji{orange}\texttwemoji{lemon} \\
    \texttwemoji{orange} Department of Computer Science \& Technology, University of Cambridge, U.K. \\
    \texttwemoji{lemon} ALTA Institute, University of Cambridge, U.K. \\
    \texttwemoji{orange} \texttt{firstname.secondname@cl.cam.ac.uk} \hspace{2mm}
}

\begin{document}
\maketitle
\begin{abstract}

Language models provide a key framework for studying linguistic theories based on prediction, but phonological analysis using large language models (LLMs) is difficult; there are few phonological benchmarks beyond English and the standard input representation used in LLMs (subwords of graphemes) is not suitable for analyzing the representation of phonemes. In this work, we demonstrate how \textbf{word segmentation} can be used as a phonological probing task, allowing us to study the representations learned by phoneme-based language models trained on child-directed speech across 31 languages. Following computational models of word segmentation, we present unsupervised methods for extracting word boundaries from a trained model using the observation that prediction-error peaks at the start of words. We also use linear probes to identify that these models implicitly track word boundaries, even when they do not appear in training.
This cross-lingual work corroborates statistical learning theories of acquisition and empirically motivates new methods for training subword tokenizers.

\begin{tblr}{colspec = {Q[c,m]|X[l,m]}, stretch = 0}
    \cincludegraphics[width=1.4em, keepaspectratio]{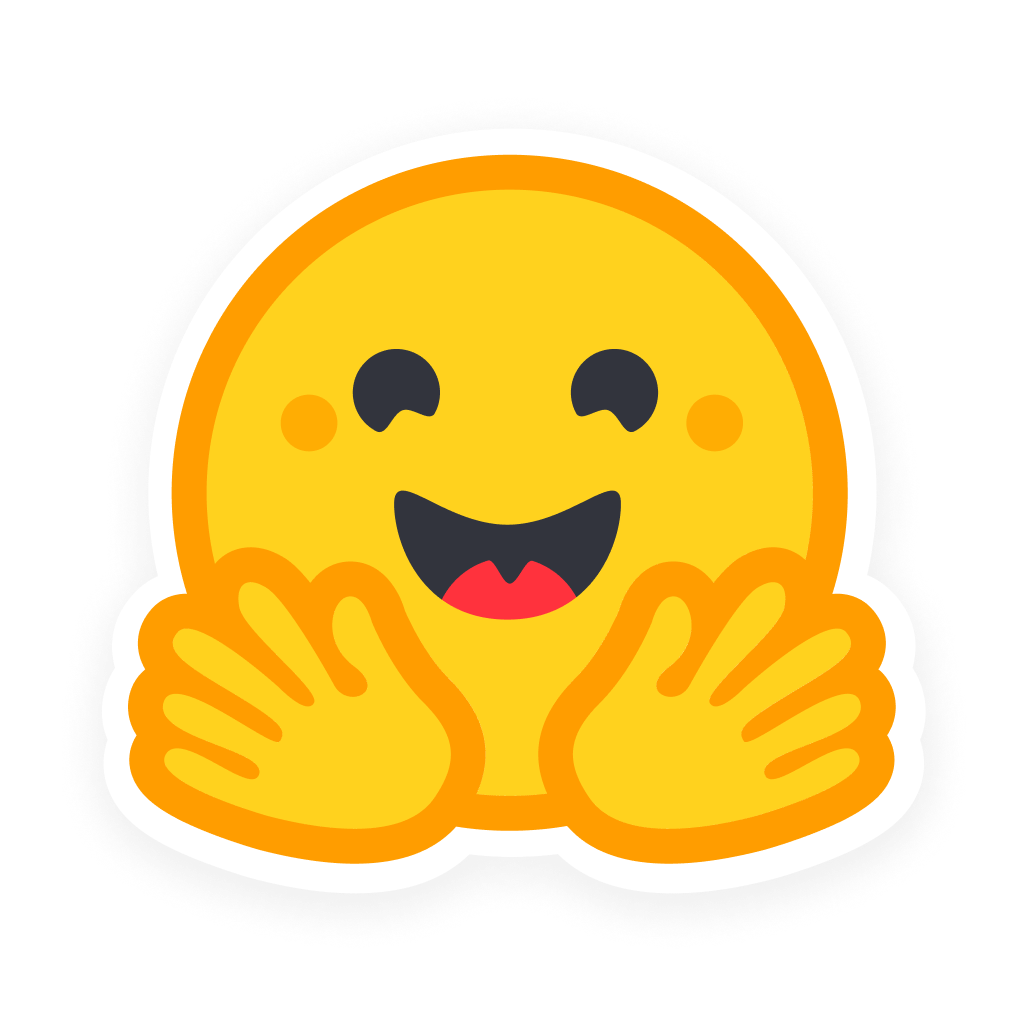} & {\footnotesize{\href{https://huggingface.co/collections/phonemetransformers/babylms-first-words-67eeadda117231f8bb055715}{phonemetransformers/first-words}}\\ \tiny{(CC BY 4.0)} } \\
    \cincludegraphics[width=1.2em, keepaspectratio]{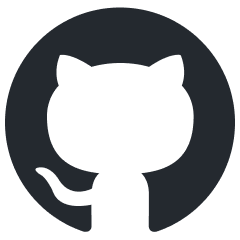} & {\footnotesize{\href{https://github.com/codebyzeb/PhonemeTransformers}{codebyzeb/phonemetransformers}}\\
    \tiny{(CC BY 4.0)}}
\end{tblr}

\end{abstract}

\vspace{2mm}

\section{Introduction}

Small models trained on developmentally plausible data have led to numerous advancements across pre-training strategies, architectures and tools for linguistic analysis \citep{hu-etal-2024-findings}. Yet most of this work involves training on English orthographic data with subword tokenization, restricting the ability to study phonological representations and word learning. A few recent studies have demonstrated that these so-called ``BabyLMs'' can be trained on individual phonemes \citep{goriely2024babble, bunzeck2024graphemes}, supporting phoneme-based phonological analysis. However, the majority of this work continues to center on English, in part due to the lack of phonological benchmarks for other languages.

In this work, we explore the phonological capabilities of phoneme-based BabyLMs across 31 languages using the \textbf{word segmentation task}. Following computational models of word segmentation studies in the acquisition literature, we investigate models by assessing their ability to correctly place word boundaries in a sequence of phonemes when word boundaries are not provided during training. Successful segmentation indicates implicit phonological knowledge and when performed zero-shot on developmentally plausible data, contributes to statistical learning theories of language acquisition. 

\begin{figure}[t]
    \centering
    \includegraphics[width=0.95\linewidth]{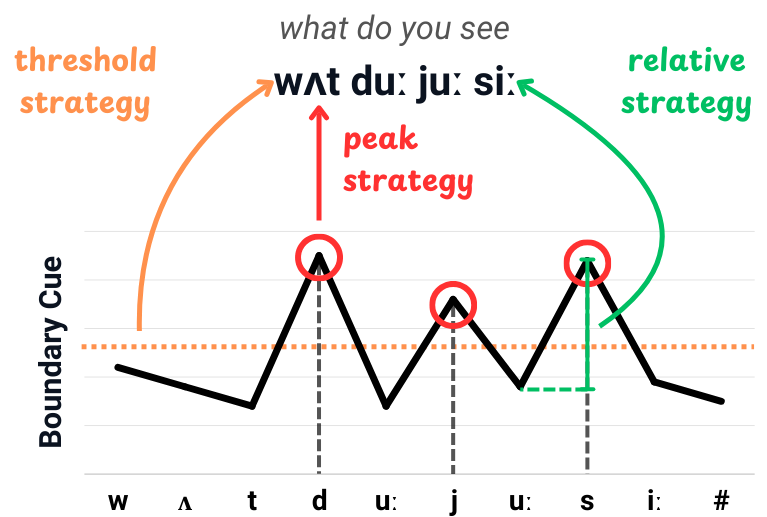}
    \caption{Three strategies for unsupervised word segmentation using cues extracted from an auto-regressive language model trained to predict phonemes.}
    \label{fig:example}
\end{figure}

In some of the earliest sequential models, it was noted that \emph{prediction-error} (the degree to which the model struggles to predict the next token) often corresponded with word boundaries \citep{elman-1990-finding}. Using this observation,
we identify four word boundary cues that can be extracted from trained models and three unsupervised strategies for placing boundaries using these cues, as illustrated in \cref{fig:example}. We additionally follow the supervised approach of \citet{hahn-baroni-2019-tabula}, training linear probes on final layer embeddings to determine if word boundaries are implicitly tracked in order to improve phoneme prediction.

We train phoneme-based BabyLMs on the phonemic transcriptions of child-centered speech comprising the \ipachildes dataset \citep{goriely2025}. We find that these models implicitly encode word boundaries across all 31 languages and identify two factors that may provide useful priors depending on the language: the length of words and the distribution of phonemes at the end of words. 

We discuss the validity of orthographic word boundaries as gold labels and note the similarities between our results and recent work that uses byte-level prediction entropy to improve the tokenization step in large language model (LLM) pre-training \citep{pagnoni2024byte}. We conclude that this framework not only supports the study of distributional phonology and acquisition, but could also have implications for improving the efficiency and robustness of LLMs.

Finally, we release our code and pre-trained models to facilitate future work.

\section{Related Work and Motivations}

Since their inception, language models have been used to study the structures of language and explore mechanisms that humans may use to learn them. 

Early ``connectionist'' language models were trained on sequences of letters or phonemes, often using developmentally plausible data in order to explore theories of word learning and phonology \citep{seidenberg1989distributed, norris1994shortlist, coltheart2001drc}. Modern \emph{large} language models (LLMs) are still probed for grammatical information, but standard benchmarks are generally based on higher-order structures: syntax and semantics rather than morphology and phonology. This is due to LLM design being optimized for downstream tasks, not linguistic analysis. For instance, LLMs are typically trained on graphemic text using subword tokens. While this representation is practical for large-scale training, these tokens are not very cognitively plausible \citep{beinborn-pinter-2023-analyzing}, are less effective than character-based tokens for learning word structure \citep{bunzeck2025subwordmodelsstruggleword} and cannot be used to explore representations of phonological units. Additionally, modern LLMs are inappropriate for theories of acquisition, due to the scales of data they are trained on \citep{warstadt-2023-babylm-findings}.

Here, we are interested in evaluating models that train directly on individual phonemes, without word boundaries. When trained on individual words, phoneme LMs have been used to study the acquisition of morphological rules \citep{kirov-2018-recurrent} and compare phonotactic complexity across languages \citep{pimentel2020phonotactic}. When trained on running text, phoneme LMs have been used for text-to-speech \citep{li-2023-phoneme-level-bert} and lyric generation \citep{ding-2024-songcomposer}. When compared to grapheme-based models on standard linguistic benchmarks, phoneme models slightly under-perform \citep{nguyen-2022-word-boundaries, bunzeck2024graphemes} but this could be attributed to pre-processing, punctuation and the fact that LLM architectures and evaluation sets have been optimized for written text \citep{goriely2024babble}. Despite the benefits of phoneme-based training, phonological evaluation is limited, and few phoneme LMs exist beyond English. \citet{goriely2025} trained phoneme LMs on child-directed speech across 11 languages, but were only able to use an English benchmark for studying how phonological and syntactic knowledge scales in phoneme LMs. 

In this work, we propose the word segmentation task as a language-independent method for probing the representations learned by phoneme LMs. Below, we summarize past approaches for investigating the phonological capabilities of language models. We then give historical background on the word segmentation task. Finally, we discuss past examples of word segmentation being used as a probing task.

\subsection{Phonological Evaluation of LLMs}

While many studies have explored the representations learned by phoneme LMs trained on individual words, there are very few benchmarks for phoneme LMs trained on running text.

One method for testing phonology is to use minimal pairs of words and pseudowords as a lexical decision task. One benchmark that uses this approach is BabySLM \citep{lavechin}, which provides a lexical decision metric for phoneme LMs or speech LMs (which learn directly from audio) using a vocabulary based on child-directed speech. \citet{bunzeck-etal-2025-small} use a similar approach in order to compare grapheme LMs to phoneme LMs. They also use two probing tasks to examine the representations of sentences; age prediction and rhyme prediction. 

PhonologyBench \citep{suvarna-etal-2024-phonologybench} is a benchmark that uses prompts to test chat-based English LLMs. However, by using prompts, they treat phonology as an emergent ability tested through metalinguistic judgment, an evaluation strategy which \citet{hu2023prompting} argues is inferior to using quantities directly derived from a model's representations.  

These benchmarks also only test English models, in part due to the lack of phoneme LMs in other languages, but also due to a lack of resources for constructing phonological tasks. For example, pseudowords are typically generated using \texttt{wuggy} \citep{keuleers2010wuggy}, which only supports three languages for phonetic pseudoword generation. An example of language-independent evaluation of phoneme LMs is the phonetic feature probe used in \citet{goriely2025}, which only requires feature vectors for each IPA symbol. The word segmentation task requires no language-specific data, only utterances labeled with word boundaries. 

\subsection{Computational Models of Segmentation}\label{sec:wordseg}

Unlike in written text, where lexical units are separated by spaces and punctuation, spoken communication consists of continuous utterances with no clear demarcation of words \citep[see, e.g.][]{cole1980model}. Somehow, without a lexicon to consult, children are able to segment speech into words and phrasal units by the age of six months \citep{Jusczyk1999infants}. How children learn to segment words and bootstrap their lexicon is known in psycholinguistics as the \emph{word segmentation problem}, and statistical learning experiments have established a wide variety of statistical cues which children may use to segment speech \citep{Cutler1987, gleitman1988learning, Jusczyk1993stress, Saffran1996distributional, Jusczyk1999allophonic, Suomi1997}.

Particularly influential were the experiments of \citet{Saffran1996learning}, who established that 8-month-old children use distributional information to segment speech, specifically noting that low conditional probability between two adjacent syllables often indicated a word boundary. These experiments inspired the development of computational models proposing cognitively plausible learning mechanisms for word segmentation, most of which are based on the principle that units within words are far more predictable than units across word boundaries \citep{harris1955}. Many models draw on \citet{Brent1999}, who use unigram statistics to segment speech, with later models using higher-order n-grams \citep{Venkataraman2001}, incorporating phonological constraints \citep{Blanchard2010} or leveraging prior distributions over word frequencies and phonological shapes \citep{Goldwater2009}. Other models explicitly calculate several statistical cues at each potential word boundary and combine cues using a majority voting framework \citep{ccoltekin2014explicit, Coltekin2017, goriely2023word}. Each cue provides a signal over the utterance (as illustrated in \cref{fig:example}) with peaks in each cue indicating a potential boundary. 

Peaks in predictability can also be observed in neural language models. In the foundational work of \citet{elman-1990-finding}, a simple recurrent network (SRN) is trained to predict letters in an unsegmented sequence (one of the first examples of auto-regressive language modeling). Elman observes that the prediction-error increases at the onset of each new word, concluding that ``there is information in the signal that could serve as a cue to the boundaries of linguistic units which must be learned''.

\citet{christiansen1998learning} later used an SRN to segment speech by using the probability of an \emph{utterance} boundary, rather than prediction-error, to place word boundaries. This followed previous work suggesting that children could use utterance boundaries to bootstrap their lexicon \citep{aslin1996models} and is a cue used in the models of \citet{ccoltekin2014explicit, goriely2023word}.

In this study, we combine ideas from past computational models for word segmentation. Rather than explicitly calculate n-gram statistics, our cues are based on prediction-error and utterance boundary probability extracted from LLMs trained on the next-phoneme prediction task. As these cues are based on the language model's prediction of phonemes, successful segmentation indicates that implicit phonological knowledge of word-like units in these models.

While our experimental setup draws on previous computational work in word segmentation, we do not claim that our phoneme-level language models simulate child language acquisition (see \cref{sec:discussion}). Rather, we use the segmentation task --- with phoneme-level input --- as a diagnostic tool that allows us to characterize the cross-linguistic distributional structure of speech sounds and test whether language models naturally group cluster sequences into units that coincide with our notion of word-hood. Although our findings may support aspects of statistical learning theories, we acknowledge the limitations of using phoneme-based representations in \cref{app:limitations}.

\subsection{Probing for Word Boundaries}

Previous work has explored the representations of word boundaries in LLMs. \citet{sanabria2021difficulty} explored methods for extracting word boundaries from attention weights in an LSTM, finding that attention had limited value for segmentation. \citet{hahn-baroni-2019-tabula} trained character-level RNNs and LSTMs without word boundaries, finding that individual activations correlated with word boundaries and that a linear probe trained on all activations also identified boundaries. They claimed that removing word boundaries resulted in a `near tabula rasa' training paradigm but trained on billions of graphemic words Wikipedia, which is not developmentally plausible. Here, we use this probe on the final layer of phoneme LMs trained on developmentally plausible data, a more `tabula rasa' paradigm. 

Other studies have verified \citeauthor{elman-1990-finding}'s observations that prediction-error corresponds with word boundaries. For instance, \citet{al2019character} train a 64-layer character-level transformer and in qualitative analysis note that three measures of prediction-error sharply increase at the start of words. However, their model is trained on graphemic text from Wikipedia without removing the word boundaries and they do not explicitly use these measures to evaluate word segmentation performance. Here, we use their three measures to propose an unsupervised word segmentation algorithm using phoneme LMs trained without word boundaries.

\section{Word Segmentation Task}

We use the \emph{word segmentation task} as a zero-shot method for studying the phonological properties of language models trained on phoneme sequences. Given a list of utterances, each of which consists of a non-delimited phoneme sequence, the task is to produce a \emph{segmentation} of each utterance by using an unsupervised method for placing word boundaries. For instance, given the utterance ``what do you see'', represented phonemically as \ttipa{w2tdu:yu:si:}, successful segmentation would return \ttipa{w2t du: yu: si:}, as demonstrated in \cref{fig:example}. Note that phonemes are individual tokens (e.g. \ttipa{u:} is a single token, not two) and, crucially, word boundaries are removed during training, although utterance boundaries are present.

Our method for unsupervised word segmentation is based on the observation made by \citet{elman-1990-finding}, that cues for word boundaries can be extracted from a sequence prediction model. Given a language model that at each position $i$ provides the probability of a phoneme $x$ given a context $x_1\ldots x_{i-1}$, we extract the following four cues at each potential boundary position:

\begin{itemize}[leftmargin=*]
    \item \textbf{Entropy:} The entropy (in bits) across the probabilities for all items in the vocabulary.
    \item \textbf{Loss:} The cross-entropy loss (bits) calculated as the negative log probability of the subsequent phoneme $p_i$.
    \item \textbf{Rank:} The rank of $x_i$ in the list of possible tokens at position $i$ sorted by likelihood.
    \item \textbf{Utterance Boundary Probability (UBP):} The probability assigned to the utterance boundary token.
\end{itemize}

The first three cues are put forward by \citet{al2019character}, where they are used to qualitatively examine the error rate of their character-based language model. Our use of these cues for word segmentation is novel. The fourth cue, UBP, relates to the model of \citet{christiansen1998learning}, who found that the prediction of the utterance boundary marker in a SRN increased at word boundaries. All four cues are utilized in the segmentation models of \citet{ccoltekin2014explicit, goriely2023word} but rather than being explicitly calculated using n-gram frequencies, we calculate them using the probability distribution produced by a language model.

For each of these cues, we have three methods for placing word boundaries. The first is to identify peaks in each cue: placing word boundaries whenever the cue's value is higher at position $i$ than at position $i-1$ or $i+1$ in the sequence. The second is to learn a single threshold value, placing word boundaries when the cue exceeds it. The third combines both strategies, placing word boundaries when the relative increase of the cue's value from position $i-1$ to $i$ exceeds a learned threshold. We call these the \textbf{peak}, \textbf{threshold} and \textbf{relative} strategies, respectively, as illustrated in \cref{fig:example}. We acknowledge that the threshold and relative strategies are not fully unsupervised, using a single learned parameter. 

Finally, in order to explore whether word boundary information is present in the model's representations, we follow \citet{hahn-baroni-2019-tabula} and train a linear probe to predict word boundaries from the final layer embeddings. We implement their `balanced' probe, training on embeddings taken from an equal number of word-final and word-internal positions, and ensure that no words in the training set are contained in the test set. 

\section{Experimental Setup}

We train a suite of GPT-2 models on each of the 31 languages in the \ipachildes corpus. As the size of each subset varies considerably,\footnote{The North American English section contains 10M words but Farsi only contains 40k.} for a fair comparison we must subsample our training data to the size of the smallest subset and use a very small model to prevent over-fitting. In order to explore the use of larger models and more training data, we train four suites of models, each using a different sample size and model size, setting model parameters according to the scaling experiments of \citet{goriely2025}. These suites are detailed in \cref{tab:suites} with parameter configurations and training parameters given in \cref{app:implementation_details}. The smallest model (only 2 layers) is trained on 100k tokens from all 31 languages, and the largest model (6 layers) is trained on 18M tokens of English. 

\setlength{\tabcolsep}{2pt}
\begin{table}[t]
    \centering
    \small
    \begin{tabular}{rccc}
    \toprule
        Suite Size & Model Parameters & Tokens (words) & Languages \\
       \midrule
       Tiny & 400k & 100k (\textasciitilde20k) & 31 \\
       Small & 600k & 700k (\textasciitilde180k) & 17 \\
       Medium & 5M & 1.8M (\textasciitilde500k) & 11 \\
       Large & 19M & 18M (\textasciitilde5M) & 1 \\
       \bottomrule
    \end{tabular}
    \caption{The model size in number of (non-embedding) parameters and data size used for each suite of models. Languages are sub-sampled according to the token count for consistency, as word length varies across languages.}
    \label{tab:suites}
\end{table}

For the linear probes, we follow \citet{hahn-baroni-2019-tabula} and report accuracy. They claim that chance performance is 50\% due to the balanced training data, but our results suggest otherwise. In order to evaluate our unsupervised strategies, we follow past work (see \cref{sec:wordseg}) compute the F1 score of boundary placement, excluding boundaries placed at the start and end of utterances (as these are `free' from the utterance boundaries).

\section{Results}

\begin{figure*}[t]
    \centering
    \includegraphics[width=\linewidth]{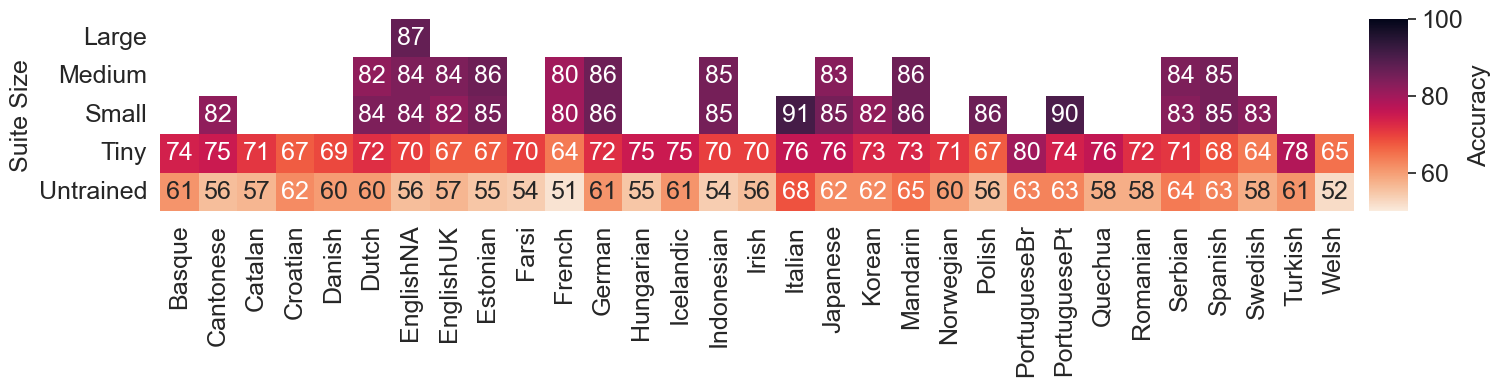}
    \caption{Accuracy scores for the word boundary probe trained on the contextual embeddings of phonemes across models in each suite. Training and test instances are balanced and each word used for training embeddings is removed from the test set. Probe results for each untrained model in the Tiny suite are included as a baseline.}
    \label{fig:probes}
\end{figure*}

\begin{figure*}[t]
    \centering
    \includegraphics[width=\linewidth]{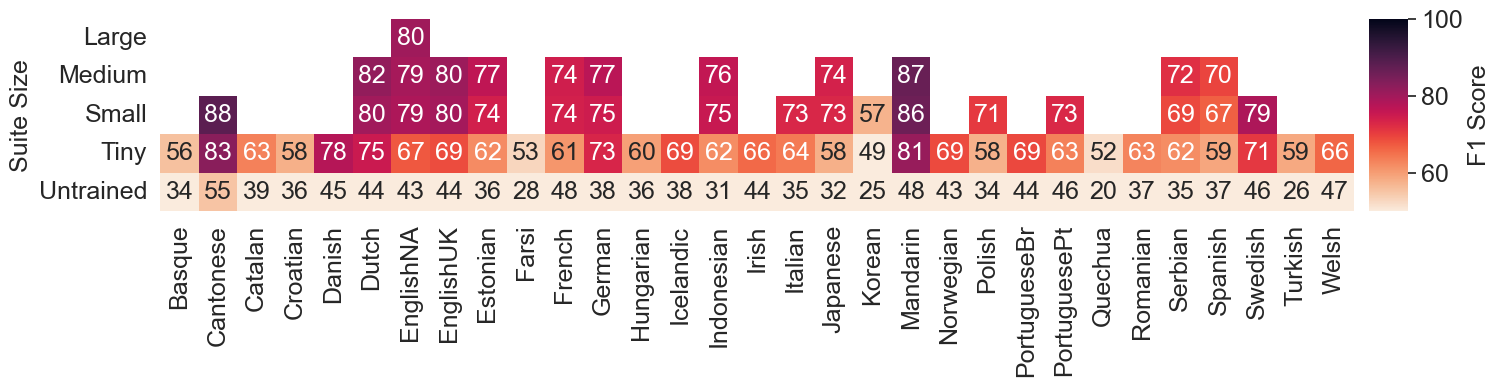}
    \caption{Boundary placement F1 scores achieved using the unsupervised segmentation strategies across models in each suite. For each score, we report the maximum across the 4 cues and 3 segmentation strategies. The Untrained row give the maximum scores achieved by each model in the Tiny suite before training.}
    \label{fig:unsupervised}
\end{figure*}

We present the results of the word boundary probe in \cref{fig:probes} and the maximum boundary F1 scores of our unsupervised segmentation strategies in \cref{fig:unsupervised}. The individual scores for each combination of language, suite size, boundary cue and segmentation strategy are provided in \cref{app:othersegresults}. 

Overall, both the word boundary probe and the unsupervised strategies successfully identify word boundaries --- all probes achieve accuracies significantly higher than the untrained baseline, as do the unsupervised strategies (see \cref{app:significance} for details on significance tests). The probe accuracies show that models implicitly track word boundaries in their contextual embeddings, suggesting that they are learning phonological rules to aid in next-phoneme prediction. The unsupervised segmentation results indicate that word boundaries can be extracted through prediction across many languages, corroborating previous statistical learning results about the role of distributional cues in language acquisition.

Below, we analyze these results in more detail.

\paragraph{180k words are sufficient for learning word boundaries.}
We note that across all languages, the accuracy of the word boundary probes increases from the Tiny suite to the Small suite (where models are trained on about 180k words, as seen in \cref{tab:suites}), but improvements are minimal for models in the larger suites. This also occurs with the unsupervised approach, despite receiving several orders of magnitude more training data and training with many more parameters. We conclude that 180k words is sufficient for a model to learn word-like units in our framework, but other models may require more or less data. 

\paragraph{Utterance boundaries are better predictors of word boundaries than prediction-error.}
\Cref{fig:unsupervised} provides the maximum boundary F1 score achieved for each model in each suite across the four boundary cues and three segmentation strategies, for a total of 12 combinations. In \cref{tab:bestcues} we summarize the cue and strategy combinations that achieved these scores. The UBP cue is the most effective in each suite, out-performing the three cues based on prediction-error, and the relative strategy out-performs the other two strategies. For reference, we give the best combinations for each language in \cref{app:othersegresults}. Generally, the best cue stays consistent across suites for a particular language (e.g. Entropy is the best cue for Italian), but this is not always the case, and the best strategy also varies. 

\begin{table}[]
    \centering
    \small
    \begin{tabular}{lcccc}
    \toprule
    Cue \& Strategy & Tiny & Small & Medium & Large \\
    \midrule
    UBP (threshold) & 3 & 2 & 1 & - \\
    UBP (relative) & 3 & 6 & 4 & - \\
    UBP (peak) & 11 & 4 & 3 & 1 \\
    Entropy (threshold) & 1 & - & 1 & - \\
    Entropy (relative) & - & 4 & 2 & - \\
    Entropy (peak) & - & 1 & - & - \\
    %Loss (threshold) & - & - & - & - \\
    Loss (relative) & 9 & - & - & - \\
    %Loss (peak) & - & - & - & - \\
    %Rank (threshold) & - & - & - & - \\
    Rank (relative) & 3 & - & - & - \\
    Rank (peak) & 1 & - & - & - \\
    \bottomrule
    \end{tabular}
\caption{Counts of the word boundary cues and segmentation strategies that achieved the highest F1 scores in each suite.}
\label{tab:bestcues}
\end{table}

\begin{figure*}
    \centering
    \includegraphics[width=\linewidth]{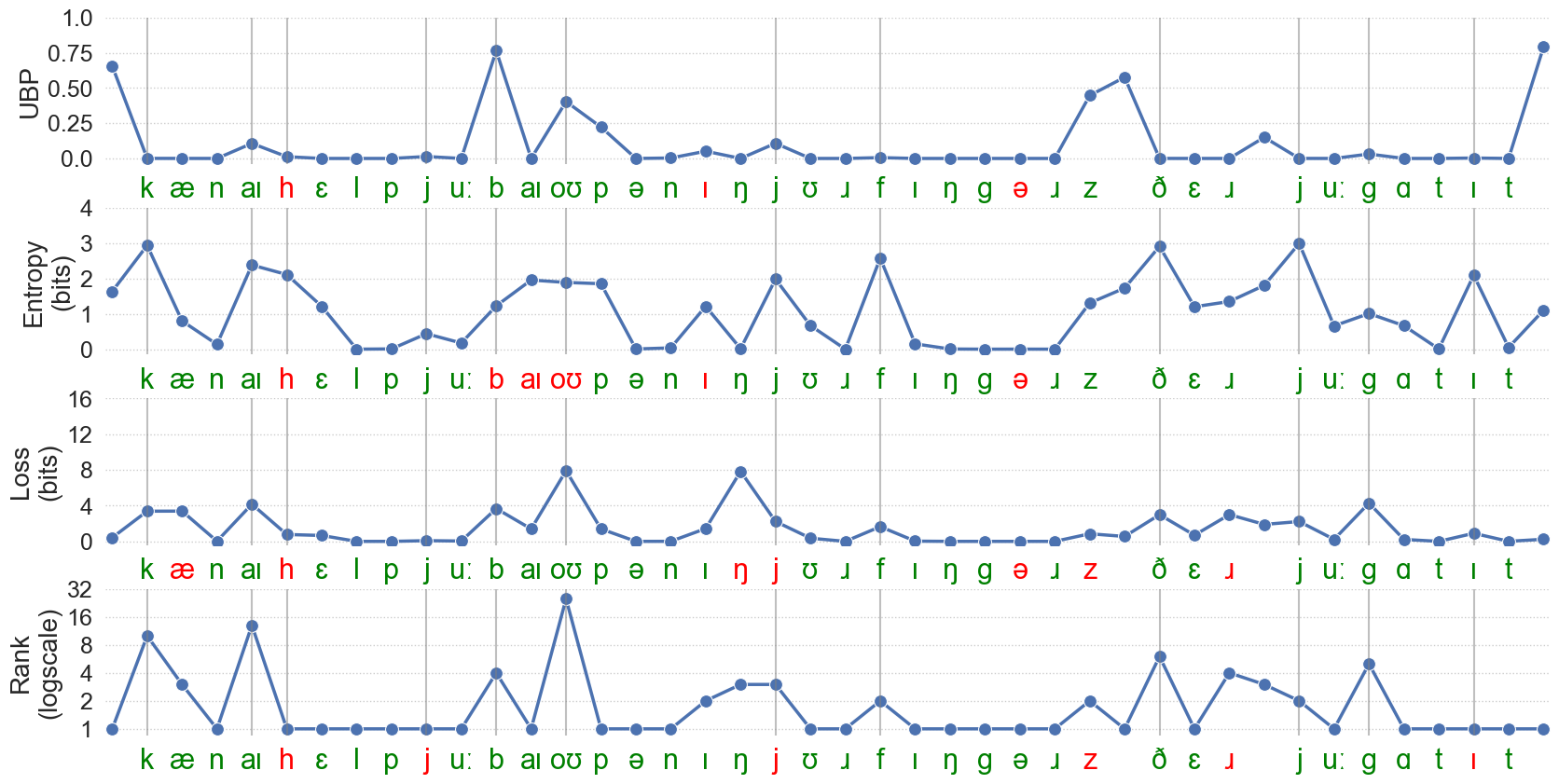}
    \caption{Per-phoneme boundary probability, entropy, loss and rank assigned by the Medium English model for the sequence of utterances ``can I help you by opening your fingers'', ``there'', ``you got it''. Spaces indicate utterance boundaries, vertical lines indicate gold word boundaries and phonemes are marked as green if they are correctly identified as word boundaries using the \textbf{peak} strategy or if they follow an utterance boundary (red otherwise).}
    \label{fig:qualitative}
\end{figure*}

\paragraph{The peak segmentation strategy fails to capture subsequent boundaries.}
We compare the four segmentation cues using the peak strategy segment utterances from the EnglishNA section of \ipachildes in \cref{fig:qualitative}. We identify two failure modes for this strategy. The first is that since two peaks cannot directly follow one another, subsequent boundaries cannot both be successfully placed. In this example, the \ttipa{h} in ``help'' is incorrectly placed by all four cues. A second failure case is that the relative size of peaks is not considered; three cues incorrectly place a boundary within the word ``fingers'' due to a very small peak at \ttipa{\textschwa}. The threshold and relative segmentation strategies address both of these issues but for English the peak strategy is still best overall.

\begin{figure}[t]
\centering
\begin{subfigure}{0.95\linewidth}
    \includegraphics[width=\linewidth]{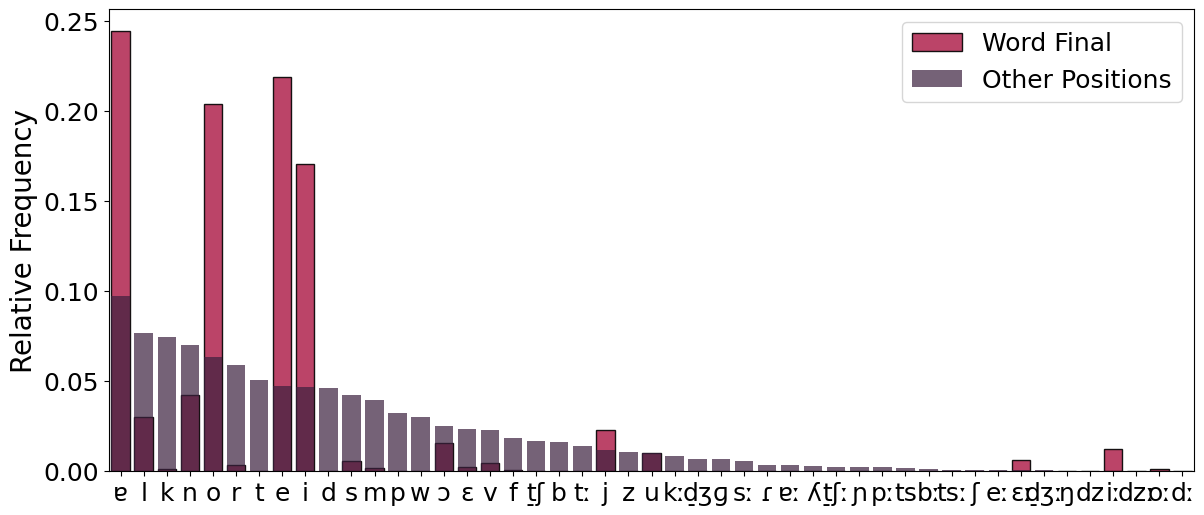}
\end{subfigure}
\hfill
\begin{subfigure}{0.95\linewidth}
    \includegraphics[width=\linewidth]{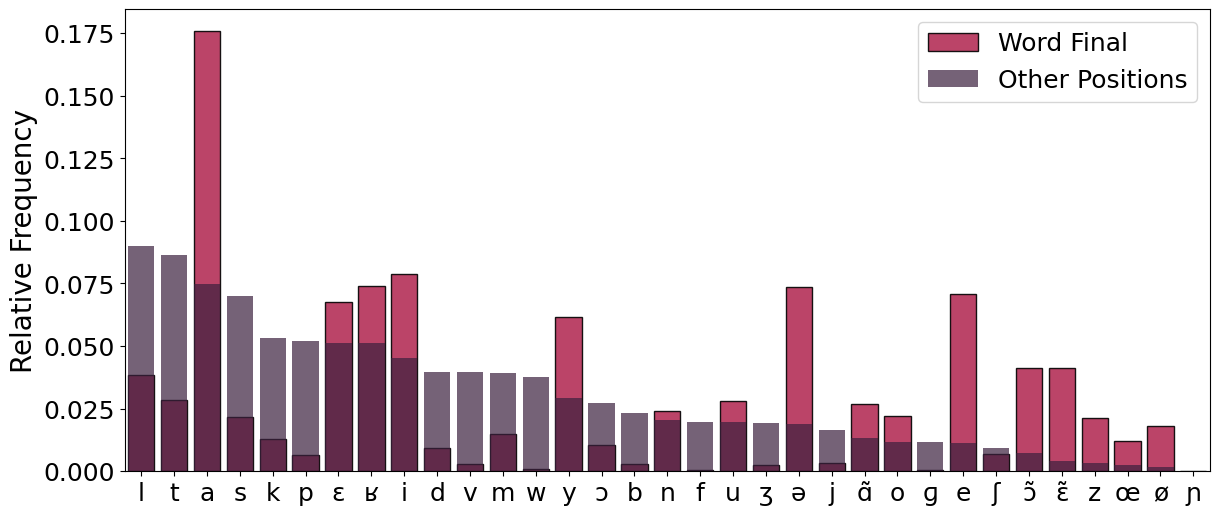}
\end{subfigure}
\hfill
\caption{Relative frequencies of phonemes appearing in word-final positions and all other positions for Italian (top) and French (bottom).}
\label{fig:frequencies}
\end{figure}

\paragraph{Italian has a strong prior for learning word boundaries.}
\citet{hahn-baroni-2019-tabula} claim that since the probes are trained on balanced examples, 
chance accuracy should be 50\%. However, we find that the probes trained on completely untrained models (see \cref{fig:probes}) achieve accuracies ranging from 51\% for French up to 68\% for Italian. This is because the balancing procedure does not account for the fact that phonemes have different probability distributions depending on their position within words. For example, in \cref{fig:frequencies} we find that at the end of Italian words, a small number of phonemes have particularly high frequencies (the vowels \ttipa{\textturna, o, e} and \ttipa{i} end 84\% of words) whereas the distribution of French word-final phonemes is not as skewed. This skewed distribution provides a useful prior for the Italian probe, which can achieve high accuracies by relying on these phoneme frequencies (the only signal available when using embeddings from an unsupervised model). To measure the relative benefit of each prior, we can compute the \textbf{normalized entropy} of the word-final phoneme distributions in each language, 

$$H_\mathrm{norm}=\frac{H(P)}{H_\mathrm{max}} = \frac{\sum_{i=1}^np_i\log_ip_i}{log_2(n)},$$ 
which ranges from 0 (deterministic distribution) to 1 (uniform distribution). We find that not only do Italian and French have the lowest and highest normalized entropies with 0.51 and 0.84, respectively, but in general, this normalized entropy has a high negative correlation with probe accuracy for the untrained models (Pearson $\rho = -0.69$). This correlation is still present for the Tiny suite (Pearson $\rho = -0.52$) but is not significant for the Small and Medium suites, indicating that although the word-final phoneme distribution prior is useful, the embeddings do still encode information about word boundaries that the probes can detect.

\paragraph{Word length is a confounding factor for unsupervised segmentation.}
Just as with the probes, using our unsupervised methods on untrained models can reveal confounding factors, as shown in \cref{fig:unsupervised}. The F1 scores for the untrained models range from 20 for Quechua up to 55 for Cantonese. For 25 of the 31 languages, this score comes from the UBP cue with the relative strategy; since the probability of an utterance boundary from an untrained model will randomly vary over the phoneme sequence, boundary placement using the relative strategy essentially places boundaries randomly, which can still yield relatively high F1 scores if words are short. This seems to be the case here; Quechua has the highest average word length in \ipachildes and Cantonese has the lowest, with 6.2 and 2.4 phonemes per word, respectively. Generally, we find that word length has a high negative correlation with the F1 scores with Pearson $\rho = -0.94, -0.71, -0.79, -0.42$ for the Untrained, Tiny, Small and Medium suites, respectively (although the final correlation is not significant). 

This confounding factor means that we cannot easily compare word segmentation scores between languages, only scores for each language across suite sizes. Compared to the untrained models, the unsupervised word segmentation strategy still achieves significantly higher F1 scores for every language, demonstrating that distributional information is a useful cue for bootstrapping a lexicon. 

\section{Discussion}\label{sec:discussion}

In this work, we train BabyLMs on phonemic transcriptions of 31 languages in \ipachildes and explore the word segmentation task as a method for probing these models for phonological knowledge. Our results indicate that prediction-error and utterance boundary probability can be used as cues for unsupervised word segmentation. Our study is the first to use prediction-error extracted from LLMs for unsupervised word segmentation, extending previous work that explicitly calculated these cues using n-gram models \citep{ccoltekin2014explicit, Coltekin2017, goriely2023word}. We also update previous neural models of word recognition \citep{elman-1990-finding, christiansen1998learning} by using modern architectures and evaluating cross-lingually. We now turn to the broader implications of our findings.

\paragraph{Statistical learning.} Viewing our models as statistical learners, we find that no single cue or strategy consistently yields the best segmentation performance across different model sizes and languages. This is perhaps unsurprising, as many of the cues are highly interrelated (for example, entropy and surprisal often correlate) and all segmentation strategies are grounded in the same underlying principle: identifying boundaries at points of high prediction uncertainty. It is this general principle, rather than any specific cue or strategy, that proves sufficient for segmenting utterances into word-like units. Nevertheless, most cues and strategies perform reasonably well on their own. Previous segmentation models have explored combining multiple distributional cues through unsupervised majority voting \citep{Coltekin2017, goriely2023word}, an approach that could be fruitfully applied to the cues investigated here in future work.

\paragraph{Cross-lingual comparison.} Comparing models across languages is a challenge. Our study is the first cross-lingual study using the word segmentation task to compare 31 languages, but we identify two confounding factors that inhibit cross-lingual comparison. Firstly, we find that the distribution of phonemes in word-final slots provides a prior not previously accounted for in studies that probed contextual embeddings for word boundary information. Secondly, we find that word length provides a prior for the unsupervised strategies, since randomly placing boundaries yields a higher F1 score when words are shorter, which has not previously been accounted for in cross-lingual word segmentation studies. Nevertheless, both the probes and the unsupervised strategies achieve significant scores for all 31 languages, indicating the importance of the distributional cue for learning to segment speech in any language. These findings also highlight the importance of accounting for frequency information as a prior when training probes or comparing models trained on different datasets.

\paragraph{Simulating acquisition.} Our results focus on the performance of our models at the end of training, whereas past work has compared the learning dynamics of phoneme-based models to developmental patterns observed in human acquisition \citep{kirov-2018-recurrent}. Although our findings indicate the utility of the distributional cue for identifying word-like units, we do not claim that our models simulate language acquisition. In particular, given recent advances in models that operate directly on raw audio, the use of phoneme-level representations may be insufficient for capturing the full complexity of language learning, as discussed in \cref{app:limitations}.

Rather, we use this framework to investigate the distributional patterns of phonemes across languages and whether language models trained to predict upcoming phonemes implicitly track meaningful sub-sequences that align with words. While many computational models of word segmentation treat segmentation as a necessary precursor for language understanding, this assumption has been questioned. For example, \citet{Baayen02012016} show that a tri-phone model, operating on unsegmented utterances can make predictions consistent with infants' sensitivity to linguistic structure. Likewise, recent phoneme-level language models perform well on both linguistic benchmarks and downstream tasks without explicit segmentation \citep{goriely2024babble} --- although our results suggest that some degree of implicit segmentation may be occurring to enhance these models' predictive performance. 

\paragraph{Word boundaries as gold labels.} Throughout this work, we have used word boundaries from orthographic text as the gold labels for evaluation, but these boundaries may not correspond with lexical units in speech. In early stages of acquisition, children may treat both predictable multi-word phrases as single lexical units \citep{macwhinney1978} and unsupervised word segmentation strategies may be segmenting morphemes, rather than words \citep{fleck2008lexicalized}. From an information-theoretic angle, word boundaries may only exist to optimize the trade-off between syntax and morphology across languages \citep{koplenig2017statistical, mosteiro2025word} and in general, what exactly defines a `word' is still up for debate \citep{dixon2002word, haspelmath2023defining}. 

\paragraph{Unsupervised segmentation for tokenization.} Instead of evaluating against word boundaries, we can treat our cues as \emph{graded} measures of co-occurrence statistics, as noted by \citet{elman-1990-finding}. This idea can be leveraged to improve the tokenization step in modern LLM pre-training. Instead of forming subwords by merging frequently occurring byte pairs, token sequences that are highly predictable can be combined. \citet{pagnoni2024byte} apply this concept to a ``token-free'' model, where bytes are joined into `patches' according to the entropy of the probability distribution for each byte (probabilities are computed using a byte-level LLM). They use two constraints for merging bytes which exactly correspond to our threshold and relative segmentation strategies, but only use entropy as a cue. In our experiments, entropy was less effective than utterance boundary probability (UBP) for unsupervised word segmentation and in an initial investigation (see \cref{app:tokenizers}) we found that creating a subword tokenizer using both cues improves the linguistic abilities of models trained on phonemes compared to regular BPE and that the UBP cue is more effective than entropy. This creates a parallel between word segmentation research and practical applications for tokenization in NLP and we encourage further work in this area.

\section{Conclusion}

Phoneme-level language models trained on developmentally plausible corpora are a valuable tool for studying cross-lingual phonology and theories of acquisition. In this study, we demonstrate how the \textbf{word segmentation task} can be used to probe these models for phonological knowledge and introduce novel unsupervised methods leveraging prediction-error and utterance boundary probability to identify words. Our findings show that models trained on 31 languages can all detect word boundaries; however, cross-linguistic comparisons are influenced by confounding factors such as word length and word-final phoneme distribution. These factors, while positing challenges, also offer new avenues for understanding the role of distributional cues in language processing cross-lingually. Finally, we explore the connection between word segmentation and information-driven tokenization schemes, highlighting how this research can inform and improve practical applications in natural language processing.

\section*{Acknowledgments}

We are grateful to Pietro Lesci and Julius Cheng for their careful reading of this article and their insightful feedback, which greatly contributed to its improvement.

Our experiments were performed using resources provided by the Cambridge Service for Data Driven Discovery (CSD3) operated by the University of Cambridge Research Computing Service, provided by Dell EMC and Intel using Tier-2 funding from the Engineering and Physical Sciences Research Council (capital grant EP/T022159/1), and DiRAC funding from the Science and Technology Facilities Council. Z\'ebulon Goriely is supported by an EPSRC DTP Studentship. 

\bibliographystyle{acl_natbib}
\bibliography{custom}

\newpage

\clearpage

\appendix

\section{Limitations}\label{app:limitations}

We acknowledge the following limitations of our work.

\paragraph{Limitations of phonemic data:} Using phonemic data for the word segmentation task is the typical framework for exploring relevant acquisition theories. However, the phonemic transcriptions in \ipachildes do have limitations. Having been generated using grapheme-to-phoneme (G2P) conversion, they may have been subject to conversion error, and the original transcriptions may also contain errors. The G2P process also removes natural variation in speech, such as accents and allophonic variation. The symbolic nature of phonemes may also be an unrealistic starting point for acquisition; it is unclear if infants have access to phonetic categories at this stage of acquisition \citep{feldman_infants_2021, mcmurray_myth_2022}. Researchers who advocate for using language models as cognitive models argue that the training data should be as developmentally plausible as possible \citep{dupoux-2018-cognitive, warstadt-2022-artificial}, and that phonemes may be as implausible as text for simulating early acquisition \citep{lavechin}.

From this perspective, a more appropriate framework is to learn segmentation directly from raw audio, as pursued in the Zero Resource Speech Challenge \citep{nguyen2020zero, dunbar2021zero}. Audio-based models naturally incorporate prosodic cues, which play a key role in language acquisition \citep{Cutler1987, Jusczyk1993stress, jusczyk-1999-stress-voice}. Unsupervised models have demonstrated the ability to perform statistical learning directly from raw speech \citep{lavechin2022can, seyssel-2023-realistic}, and have found that the resulting units tend to be shorter than phonemes, consistent with early perceptual categories \citep{schatz2021early}. While such models show promising signs of early phonetic learning and perform well on word-level tasks, they currently require significantly more data to match the performance of text-based models \citep{lavechin}. Moreover, training on curated audiobook datasets gives these models a considerable advantage over learning from noisier, long-form audio that better resembles real-world input—but ongoing work is making such realistic simulations increasingly viable \citep{lavechin2024modeling}.

\paragraph{Distribution of languages:} When training models cross-lingually, we were limited by the scale of each language partition of the \ipachildes dataset. The dataset has a very skewed distribution: the EnglishNA section contains 18M words but the Farsi section only contains 43k words. We addressed this skew by training four suites of models in order to provide a cross-lingual comparison while also exploring how segmentation performance increased in scale for the languages with more data available.

\paragraph{Language coverage:}
To the best of our knowledge, our work is the most cross-lingual exploration word segmentation to date, but is still limited in language coverage: the languages we compare are predominantly European and Asian, with no languages indigenous to the Americas, Australia or Africa. Word segmentation of languages that are more globally distributed should be explored in future work.

\section{Implementation Details}\label{app:implementation_details}

We conduct our experiments using the \texttt{PyTorch} framework \citep{paszke-etal-2019-pytorch} and the \texttt{Transformers} library \citep{wolf-etal-2020-transformers}.

\subsection{Hardware Details}

We use a server with one NVIDIA A100 80GB PCIe GPU, 32 CPUs, and 32 GB of RAM for all experiments. Below, we report a subset of the output of the \emph{lscpu} command:

\begin{tcolorbox}[left=5pt,right=5pt,top=5pt,bottom=5pt]
\small
\begin{verbatim}
Architecture:        x86_64
CPU op-mode(s):      32-bit, 64-bit
Address sizes:       46 bits physical, 
                     48 bits virtual
Byte Order:          Little Endian
CPU(s):              32
On-line CPU(s) list: 0-31
Vendor ID:           GenuineIntel
Model name:          Intel(R) Xeon(R)
                     Silver 4210R CPU
                     @ 2.40GHz
CPU family:          6
Model:               85
Thread(s) per core:  1
Core(s) per socket:  1
Socket(s):           8
Stepping:            7
BogoMIPS:            4800.11
\end{verbatim}
\end{tcolorbox}

\subsection{Model Parameters and Training Procedure}

\begin{table}[h!]
    \centering
    \small
    \begin{tabular}{lcccc}
    \toprule
         Parameter & Tiny & Small & Medium & Large \\
    \midrule
         Layers & 2 & 3 & 6 & 6\\
         Heads & 4 & 4 & 8 & 8 \\
         Dropout & 0.3 & 0.3 & 0.3 & 0.1 \\
         Embedding Size & 128 & 128 & 256 & 512 \\
         Inner Size & 512 & 512 & 1024 & 2048 \\
         \midrule
         Max Example Length & \multicolumn{4}{c}{128} \\
         Learning Rate & \multicolumn{4}{c}{0.001}\\
         Optimizer & \multicolumn{4}{c}{AdamW} \\
         Scheduler Type & \multicolumn{4}{c}{Linear}\\
         Max Steps & \multicolumn{4}{c}{200k} \\
         Warm-up Steps & \multicolumn{4}{c}{60k} \\
         Per Device Batch Size & \multicolumn{4}{c}{32} \\
    \bottomrule
    \end{tabular}
    \caption{Hyperparameter settings for training the GPT-2 architecture in each suite. Vocabulary size varies according to the language, but all other parameters are constant across experiments. Where values are not reported, they may be assumed to be default values.}
    \label{table:baseline_hyperparams}
\end{table}

We describe the model and training parameters in \cref{table:baseline_hyperparams}. The model parameters were chosen according to the scaling experiments of \citet{goriely2025}, who trained a suite of GPT-2 models for different subsets of the English section of \ipachildes and used the lexical score in BabySLM \citep{lavechin} to determine the best parameters. We note that since these parameters were optimised for English, there may be better parameters for the other languages, but differences in perplexity between languages were generally larger than the differences in perplexity between models in the scaling experiments we reference.

Data is prepared into batches by first tokenizing the entire dataset, combining all tokens into one long vector, and then splitting the vector into chunks of 128 tokens. Only the very last example is padded, if required. At each step during training, random chunks are selected and combined into batches. 

Checkpoints are taken every 20,000 steps during training. At each checkpoint, the perplexity is evaluated on the held-back evaluation set, and at the end of training the checkpoint with the lowest perplexity is returned as the best model. For the Tiny suite, many of the best models were from the very first checkpoint, since due to the small training dataset and small model, the model had already fit the data by this point.

\section{Full Word Segmentation Results}\label{app:othersegresults}

\begin{figure*}
    \centering
    \includegraphics[width=0.99\linewidth]{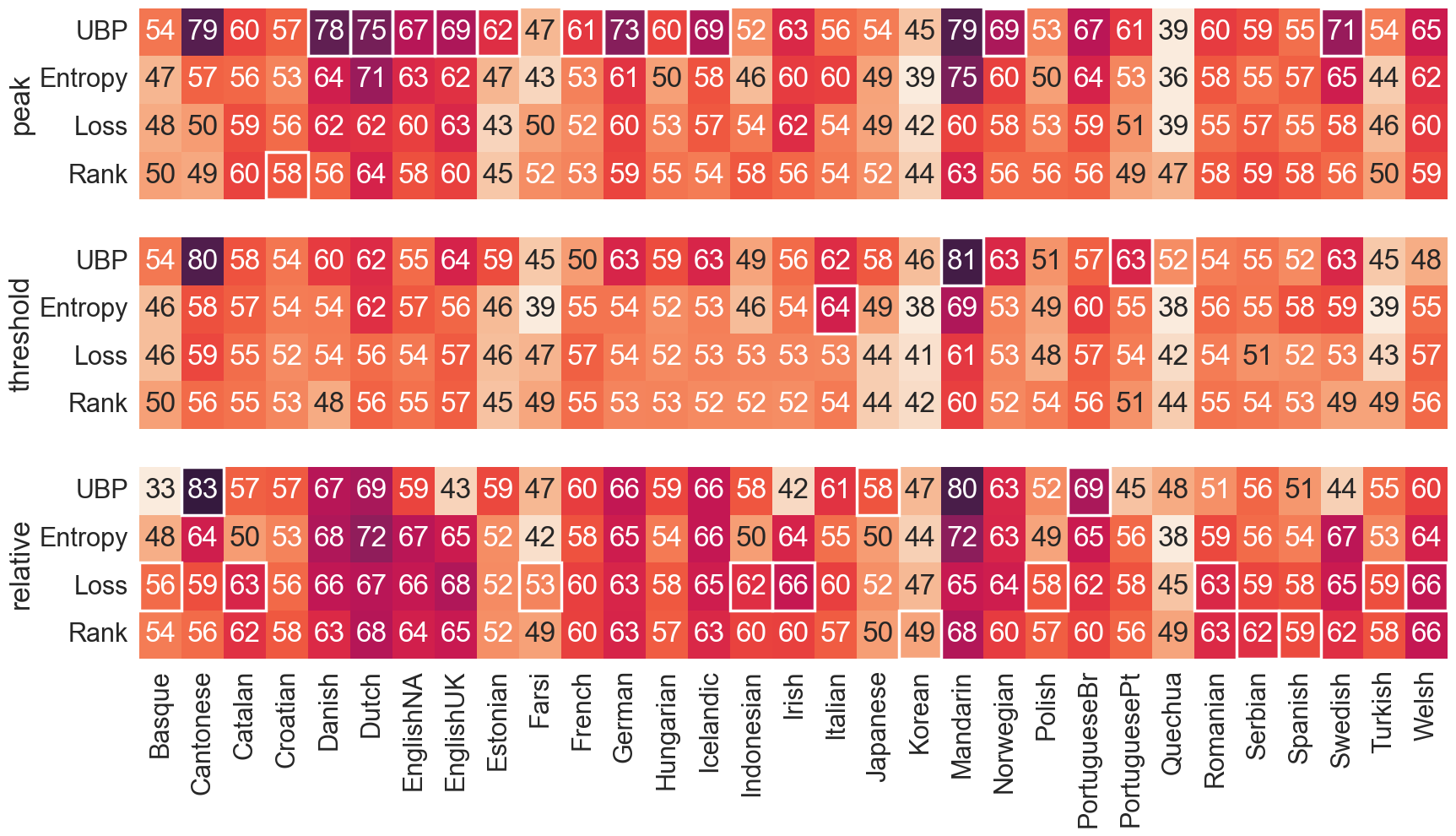}
    \caption{Boundary placement F1 scores achieved by the models in the \textbf{Tiny} suite for each cue and segmentation strategy, with the highest score for each language highlighted.}
    \label{fig:tiny}
\end{figure*}

\begin{figure*}
    \centering
    \includegraphics[width=0.99\linewidth]{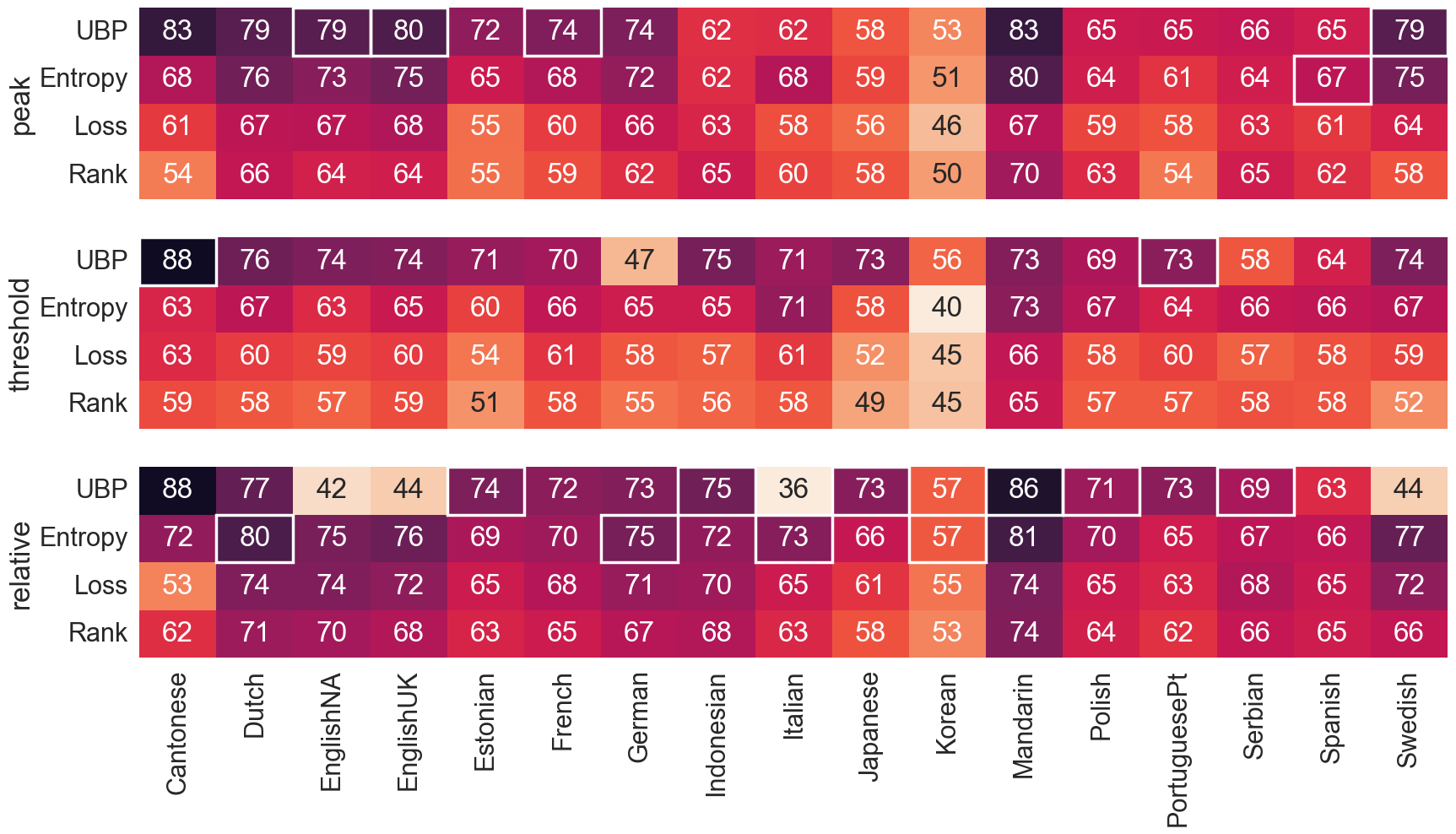}
    \caption{Boundary placement F1 scores achieved by the models in the \textbf{Small} suite for each cue and segmentation strategy, with the highest score for each language highlighted.}
    \label{fig:small}
\end{figure*}

\begin{figure*}
    \centering
    \includegraphics[width=0.99\linewidth]{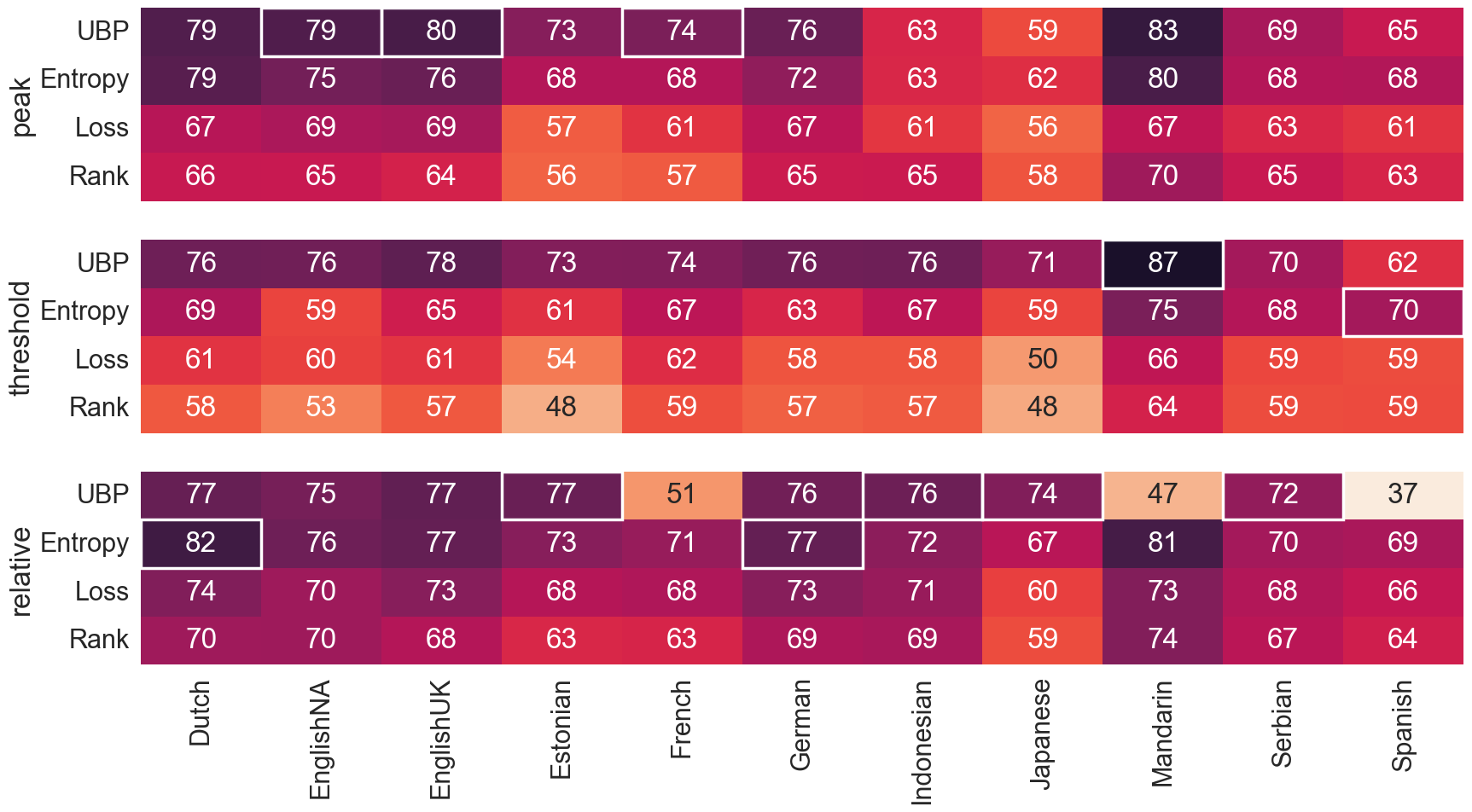}
    \caption{Boundary placement F1 scores achieved by the models in the \textbf{Medium} suite for each cue and segmentation strategy, with the highest score for each language highlighted.}
    \label{fig:medium}
\end{figure*}

\begin{figure*}
    \centering
    \includegraphics[width=0.99\linewidth]{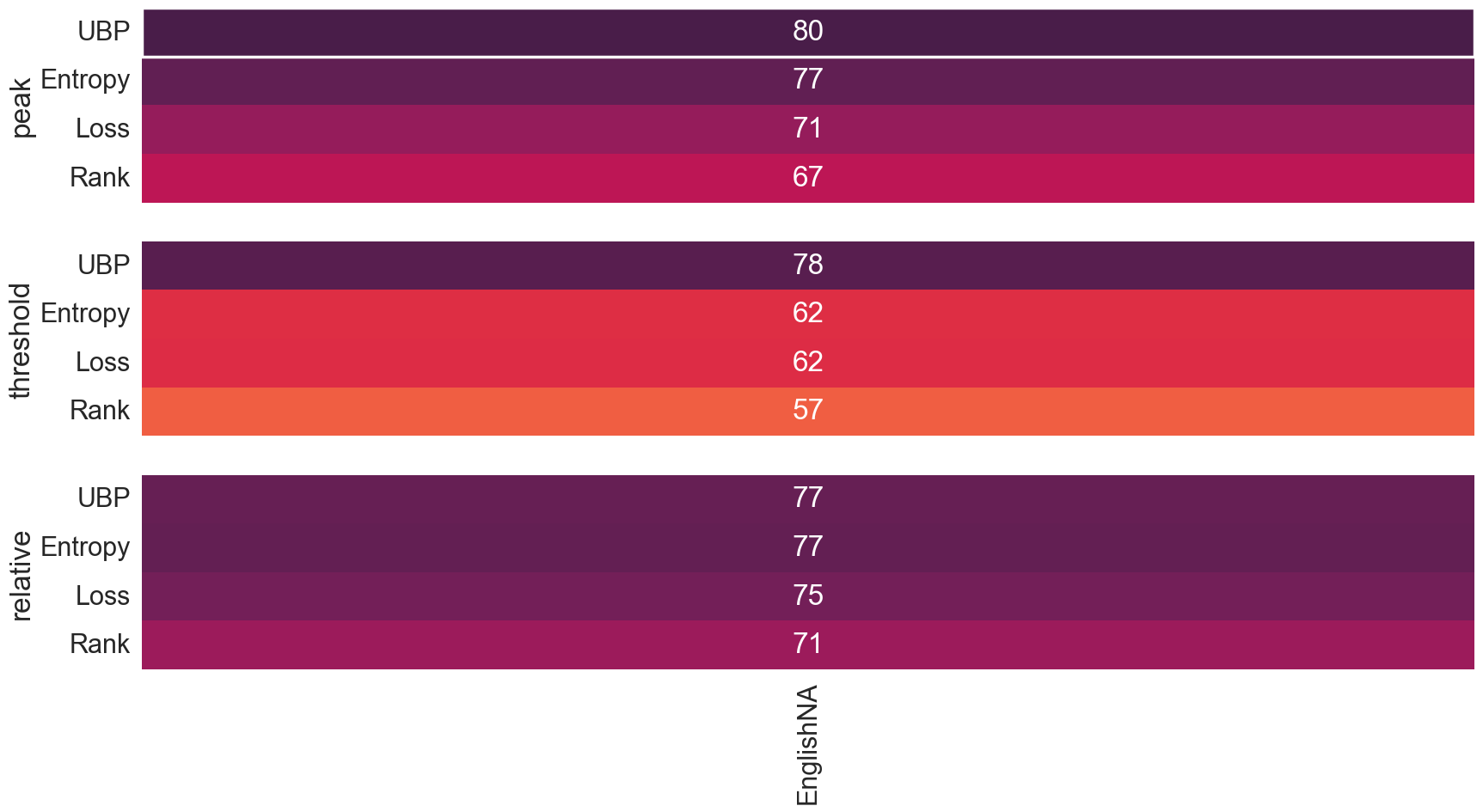}
    \caption{Boundary placement F1 scores achieved by the models in the \textbf{Large} suite for each cue and segmentation strategy, with the highest score for each language highlighted.}
    \label{fig:large}
\end{figure*}

All boundary placement F1 scores for the Tiny, Small, Medium and Large suites are given in \cref{fig:tiny}, \cref{fig:small}, \cref{fig:medium} and \cref{fig:large}, respectively. The best combination of cue and segmentation strategy for each language is given in \cref{tab:bestcuesfull}.

\begin{table*}[t]
    \centering
    \small
    \begin{tabular}{lllll}
    \toprule
    Language & 100k & 700k & 2M & 18M \\
    \midrule
    Basque & Loss (relative) &  &  &  \\
    Cantonese & UBP (relative) & UBP (threshold) &  &  \\
    Catalan & Loss (relative) &  &  &  \\
    Croatian & Rank (peak) &  &  &  \\
    Danish & UBP (peak) &  &  &  \\
    Dutch & UBP (peak) & Entropy (relative) & Entropy (relative) &  \\
    EnglishNA & UBP (peak) & UBP (peak) & UBP (peak) & UBP (peak) \\
    EnglishUK & UBP (peak) & UBP (peak) & UBP (peak) &  \\
    Estonian & UBP (peak) & UBP (relative) & UBP (relative) &  \\
    Farsi & Loss (relative) &  &  &  \\
    French & UBP (peak) & UBP (peak) & UBP (peak) &  \\
    German & UBP (peak) & Entropy (relative) & Entropy (relative) &  \\
    Hungarian & UBP (peak) &  &  &  \\
    Icelandic & UBP (peak) &  &  &  \\
    Indonesian & Loss (relative) & UBP (relative) & UBP (relative) &  \\
    Irish & Loss (relative) &  &  &  \\
    Italian & Entropy (threshold) & Entropy (relative) &  &  \\
    Japanese & UBP (relative) & UBP (relative) & UBP (relative) &  \\
    Korean & Rank (relative) & Entropy (relative) &  &  \\
    Mandarin & UBP (threshold) & UBP (relative) & UBP (threshold) &  \\
    Norwegian & UBP (peak) &  &  &  \\
    Polish & Loss (relative) & UBP (relative) &  &  \\
    PortugueseBr & UBP (relative) &  &  &  \\
    PortuguesePt & UBP (threshold) & UBP (threshold) &  &  \\
    Quechua & UBP (threshold) &  &  &  \\
    Romanian & Loss (relative) &  &  &  \\
    Serbian & Rank (relative) & UBP (relative) & UBP (relative) &  \\
    Spanish & Rank (relative) & Entropy (peak) & Entropy (threshold) &  \\
    Swedish & UBP (peak) & UBP (peak) &  &  \\
    Turkish & Loss (relative) &  &  &  \\
    Welsh & Loss (relative) &  &  &  \\
    \bottomrule
    \end{tabular}
    \caption{Best combination of boundary cue and segmentation strategy for each language and each suite.}
    \label{tab:bestcuesfull}
\end{table*}

\section{Significance Tests}\label{app:significance}

All word boundary probes for a particular language are trained and tested on the same evaluation set. We compute significance between two probes using McNemar’s Test \citep{McNemar_1947} over the predicted word boundaries for the evaluation set, with a significance threshold of $p<0.05$. The same procedure is used when comparing the unsupervised methods.

\section{Using Word Segmentation Cues for Subword Tokenization}\label{app:tokenizers}

We briefly explore the use of our unsupervised word boundary cues to create a subword tokenizer. Typically, the vocabularies for these tokenizers are generated using methods like Byte-Pair Encoding \citep{sennrich-etal-2016-bpe}, where the vocabulary initially consists of each individual byte, and pairs of bytes that frequently co-occur in a training dataset are `merged' into a new token, with this process repeated until a fixed vocabulary size is reached. We use the same principle, but base merges on the word boundary cues from a language model trained on the dataset.

Our method is as follows:

\begin{enumerate}
    \item We take a trained phoneme-level LM and compute either the UBP cue or the entropy cue at every position in the a given dataset. 
    \item We initialize our vocabulary $V$ to match the vocabulary of the phoneme LM (so it contains every phoneme plus the utterance boundary token).
    \item For every pair of tokens $x_i, x_j \in V$ that co-occur in the dataset, we compute the score for that pair by finding the average value of the word boundary cue at the position of the second token in the pair (e.g. for the pair \ttipa{D,E}, we find the value of the cue at every position where \ttipa{E} appears after \ttipa{D} and return the average). 
    \item We find the pair with the lowest score, create a new token $V_i+V_j$, add it to the vocabulary and apply the merge to every token in the dataset. The cue's value at the newly merged token is set to be the sum of the cue's value of the two tokens before the merging occurs. For the entropy cue this follows from the chain rule and for the UBP cue this results in the probability that \emph{either} original token was an utterance boundary.
    \item We repeat (2)-(3), adding new tokens and applying merges until a fixed vocabulary size is reached.
\end{enumerate}

Conceptually, creating merges using minimum average entropy will join highly predictable tokens together and result in tokens with comparable information and a uniformly dense signal that the model can learn from. Creating merges using the minimum average probability of an utterance boundary is similar, but instead tokens are joined according to the model's certainty that they do not cross an utterance boundary. 

In order to test this method, we use the phoneme-level LM trained by \citet{goriely2024babble} on a phonemized version of the 100-million word BabyLM dataset \citep{choshen-et-al-2024-callforpapers-babylm2} and train subword tokenizers using a phonemized version of the 10-million word BabyLM dataset. We create two tokenizers with a vocabulary size of 16k using the UBP cue and the entropy cue. We compare these to the BPE tokenizer trained by \citet{goriely2024babble} on the same dataset, which also has a vocabulary size of 16k. Note that all three tokenizers are trained on a dataset without word boundaries, so it is possible for tokens to span word boundaries.

\citet{goriely2024babble} trained a large model using their BPE tokenizer on the 100-million word BabyLM dataset and evaluated their results on two linguistic benchmarks, BLIMP \citep{warstadt-2020-blimp} and BabySLM \citep{lavechin}. We train and evaluate a model using the same procedure but replace their tokenizer for ours. 

\begin{table}[t]
    \centering
    \begin{tabular}{lccc}
        \toprule
        Tokenizer & \rotatebox[origin=l]{90}{BLIMP} & \rotatebox[origin=l]{90}{BabySLM Syntactic} & \rotatebox[origin=l]{90}{BabySLM Lexical} \\
        \midrule
        BPE & 71.7 & 74.7 & 71.2 \\
        Entropy & 72.7 & 77.6 & 81.3 \\
        UBP & 72.6 & 85.6 & 84.4 \\
        \bottomrule
    \end{tabular}
    \caption{BLIMP and BabySLM scores achieved by a GPT-2 model trained on the BabyLM dataset. We compare BPE to our subword method, where merges are assigned using either entropy or UBP as a cue. BPE results are taken from \citet{goriely2024babble}.}
    \label{tab:tokenizerresults}
\end{table}

The results of this experiment are provided in \cref{tab:tokenizerresults}. We find that our two tokenizers improve all three scores compared to the BPE tbut instead okenizer with the UBP cue leading to a particularly large improvement for the BabySLM syntactic score.

Our method is similar to \citet{pagnoni2024byte}, who calculate the entropy cue over bytes using a small byte-level LLM, and use either a \textit{global constraint} (corresponding to our threshold segmentation strategy) or a \textit{monotonic constraint} (corresponding to our relative segmentation strategy) in order to group bytes into latent `patches'. These patches are then fed into the main model, a large transformer, and the encoded patches are `unpatched' and fed back into the byte-level LLM to predict the next byte. Future work should investigate whether their method is improved by using the cues explored in this study. When training with word boundaries, the prediction of the space character (or other word boundary characters) could also be used to group bytes.

\end{document}